# Executable Ontologies in Game Development: From Algorithmic Control to Semantic World Modeling

Alexander Boldachev

boldachev@gmail.com

https://orcid.org/0000-0002-7259-2952

Abstract

This paper examines the application of Executable Ontologies (EO), implemented through the boldsea framework, to game development. We argue that EO represents a paradigm shift: a transition from algorithmic behavior programming to semantic world modeling, where agent behavior emerges naturally from declarative domain rules rather than being explicitly coded. Using a survival game scenario (*Winter Feast*), we demonstrate how EO achieves priority-based task interruption through dataflow conditions rather than explicit preemption logic. Comparison with Behavior Trees (BT) and Goal-Oriented Action Planning (GOAP) reveals that while these approaches model *what agents should do*, EO models *when actions become possible* - a fundamental difference that addresses the semantic-process gap in game AI architecture. We discuss integration strategies, debugging advantages inherent to temporal event graphs, and the potential for LLM-driven runtime model generation.

## 1. Introduction

### 1.1. The Crisis of Imperative Control in Complex Game Systems

The development of complex game world simulation systems has reached a point of architectural inflection. The exponential growth in simulation complexity - from persistent Massively Multiplayer Online games (MMO) to open worlds with thousands of non-player characters (NPC) - has exposed a fundamental crisis of imperative control, which becomes increasingly difficult to maintain and extend. Traditional approaches based on algorithmic control produce a semantic gap: a structural mismatch between declarative knowledge about the world (rules, lore) and its procedural implementation in code.

Behavior Trees (BT), the current industry standard, excel at encoding predictable behavioral sequences. Finite State Machines (FSM) provide clear state-transition logic. Goal-Oriented Action Planning (GOAP) offers dynamic plan generation. Yet all three approaches share a common limitation: they model behavior as algorithms - explicit sequences of decisions and actions that must be programmed, tested, and maintained separately from the semantic structure of the game world.

The crisis of imperative control in game development manifests in three fundamental problems. First, the *semantic-process gap*: declarative knowledge about the world becomes fragmented and scattered across various procedural codebases, hindering scalability and maintenance. Second, *performance limitations* due to constant polling (in BT), which inefficiently consumes CPU resources checking the state of thousands of idle agents. Third, *structural rigidity* of control structures, requiring complex manual refactoring and restructuring of behavioral logic (for example, to introduce a new priority level), making systems brittle and difficult to scale in late development stages.

## 1.2. The Core Thesis: Behavior as Emergent Property

This paper proposes an approach to overcoming the described crisis based on Executable Ontologies (EO), implemented in the boldsea framework (Boldachev, 2025a)[1]. EO is considered not merely as an alternative method of imperative control, but as a transition to a different level of abstraction: semantic modeling of the game world instead of algorithmic control.

| Approach | Designer specifies | Agent behavior is |
|---|---|---|
| BT/FSM | Decision tree / state graph | Executed algorithm |
| GOAP | Actions + preconditions + goals | Generated plan |
| EO | Event-based world ontology | Consequence of condition satisfaction |

*Table 1. Designer control of agent behavior.*

In the EO paradigm, the designer does not explicitly program behavior. Instead, an event-semantic model of the domain is created: what entities exist, what properties they possess, and under what conditions various events can occur.

## 1.3. Contribution and Structure

This paper makes three contributions:

*Conceptual reframing.* We propose EO as a foundation for transitioning from programming the behavior of individual elements to modeling executable worlds.

*Practical demonstration.* Through the Winter Feast survival scenario, we show how priority-based task interruption emerges from condition logic without explicit preemption mechanisms.

*Integration analysis.* We examine how EO can complement existing game architectures, serving as a high-level semantic layer above Entity Component System (ECS) and physics systems.

The remainder of this paper is organized as follows. Section 2 analyzes the semantic-process gap in current game AI approaches. Section 3 presents EO core principles. Section 4 details the Winter Feast implementation. Section 5 discusses integration strategies. Section 6 provides comparative analysis. Section 7 cover architectural properties and limitations. Section 8 concludes.

---

[1] In this paper we use the term Executable Ontologies (EO) in two related senses. At the conceptual level, EO denotes a general paradigm in which semantic models are directly executable. At the implementation level, all examples are instantiated using the boldsea framework as a concrete event-driven realization of this paradigm.

# 2. The Semantic-Process Gap in Game AI

## 2.1. Knowledge Representation vs. Behavior Execution

Game development inherently involves two distinct activities: defining what the game world is (ontology) and specifying how entities behave (control). In current practice, these activities use different representations, different tools, and often involve different team members.

The design document contains semantic knowledge: "A survivor in the wilderness must stay warm and fed to remain safe. When cold, the survivor should prioritize finding warmth over searching for food - you cannot hunt effectively while freezing. Gathering wood and lighting a fire restores warmth; hunting, cooking, and eating restore energy." This is declarative, readable, and captures the designer's intent.

The implementation fragments this unified knowledge across multiple systems: resource managers track thresholds, behavior trees encode priority through structural position rather than semantic meaning, action preconditions are defined separately from the survival logic that motivates them. The original declarative knowledge has been compiled into imperative code.

This fragmentation creates the *semantic-process gap*: the structural disconnection between what designers know about the game world and how that knowledge is operationalized in code (Tutenel, et al., 2009, Kessing, 2012).

## 2.2. Behavior Trees: Algorithms Detached from Semantics

Behavior Trees organize agent decision-making as hierarchical trees of nodes. Control nodes (Sequence, Fallback, Parallel) manage execution flow; leaf nodes perform actions or check conditions. A "tick" signal propagates through the tree each frame, determining which action to execute.

Strengths:

- Intuitive visual representation
- Modular composition of behaviors
- Reactivity through priority ordering (left branches first)

Semantic limitations:

- The tree encodes control flow, not world knowledge
- Conditions are evaluated procedurally, not derived from semantic state
- Adding a new rule (e.g., "fear of fire") requires restructuring the tree
- No inherent connection between the behavior and the domain model

In BT, the world is accessed through a "blackboard" - an external key-value store that the tree queries. The behavior has no semantic understanding of what the blackboard values mean; it simply checks thresholds and executes corresponding actions.

## 2.3. GOAP: Planning Over Flat State

Goal-Oriented Action Planning represents agent capabilities as operators with preconditions and effects. Given a goal state, a planner (typically A*) searches for a sequence of actions that transforms the current world state into the goal state.

Strengths:

- Dynamic plan generation - adapts to world changes
- Clear separation of goals from methods
- Emergent behavior from action composition

Semantic limitations:

- World state is a flat key-value snapshot, not a structured model
- No temporal history - only current state matters for planning
- Preconditions are Boolean predicates, not semantic relationships
- Plan execution is separate from plan validation

GOAP asks: "Given these actions and this goal, what sequence achieves the goal?" It does not ask: "Given this world model, what events are currently possible?" The difference is between searching for a solution and recognizing opportunity.

## 2.4. The Alternative: Modeling the World, Not the Behavior

Executable Ontologies propose a different question: instead of "How should the agent behave?", ask "What is true about this world, and what becomes possible when conditions are met?" (Lapeyrade, 2022).

Consider the survival scenario: an agent must maintain warmth and energy. The imperative approach programs a behavior tree that checks warmth, decides to gather wood, then light fire, then warm up. The semantic approach models:

- Agents have warmth and energy properties
- `warmthLow` is true when `warmth < warmthMin`
- Action `gather_wood` is possible when
  `warmthLow == 1 && hasWood == 0 && location.hasTree == 1`
- When fire exists and agent has wood and is cold, warming occurs

No "decision" is programmed. The agent gathers wood because, at that moment in the semantic graph, the condition for gathering wood becomes true. If warmth drops during hunting, the hunting action's condition (`warmthLow == 0`) becomes false, and it is no longer available. The "interruption" is not a programmed preemption - it is a consequence of the world state changing.

This is the core insight of EO for game development: behavior is not coded; it emerges from the ontology.

# 3. Executable Ontologies: Core Principles

## 3.1. Event Semantics: The World as Temporal Graph

The theoretical foundation of EO is the Subject-Event Ontology (Boldachev, 2025b), which reconceptualizes activity not as object state changes but as a graph of semantically typed events linked by causal relationships.

Each event follows a unified structure:

```
Event = (Id, Base:Type:Value, Actor, Cause, Model, Timestamp)
```

Where:

- `Base`. The event to which this event refers, including initiation events of entity and its properties (enabling properties-on-properties)
- `Type`. The property being recorded (attribute, relation, or role)
- `Value`. The specific value
- `Actor`. Who recorded this event (player, NPC controller, sensor, system)
- `Cause`. Links to prerequisite events (forming the causal graph)
- `Model`. The template that validated this event's creation
- `Timestamp`. For UI/audit purposes (not for ordering)

*Key distinction from traditional game state.* In ECS or blackboard systems, state is a mutable snapshot - values are overwritten, history is lost. In EO, the event graph is append-only and immutable. The "current state" is computed by traversing the graph, but all history remains accessible. This provides native support for:

- *Replay systems* - the graph *is* the replay
- *Debugging* - trace any value to its causal origins
- *Undo/save systems* - restore by truncating or branching the graph
- *Multiplayer reconciliation* - merge graphs from different actors

## 3.2. Two-Level Structure: Models and Reification Events

EO maintains a strict separation between schema and instance, analogous to TBox/ABox in description logics:

*Model Events* define templates - what properties a concept can have, what conditions must be met, what restrictions apply[2]:

```
Survivor: Model: Model Survivor
 : Attribute: warmth
```

---

[2] Examples code uses the native EO notation - Boldsea Semantic Language (BSL).

```
  : Attribute: warmthLow
  :: SetValue: +$.warmth < +$.warmthMin
  : Attribute: action_gather
  :: Condition: $.warmthLow == 1 && $.hasWood == 0
  :: SetDo: …
```

*Reification Events* are concrete instances created according to models:

```
  Survivor: Individual: John Doe
  : warmth: 70
  : hasWood: 0
```

The critical constraint: *reification events can only be created if they satisfy their model's conditions and restrictions*. This provides runtime validation - invalid states cannot exist because the engine refuses to create events that violate model constraints.

## 3.3. Dataflow Execution: Activation by Data Readiness

The fundamental difference between EO and control-flow architectures lies in the execution model[3].

*Control-flow (BT/FSM).* A tick signal propagates through the structure each frame. Every condition is polled. The system asks: "What should I do now?" even when nothing has changed.

*Dataflow (EO).* Events are activated when their conditions become true. The engine maintains subscriptions. When a new event is written to the graph, dependent conditions are re-evaluated. No polling - only reaction to actual changes.

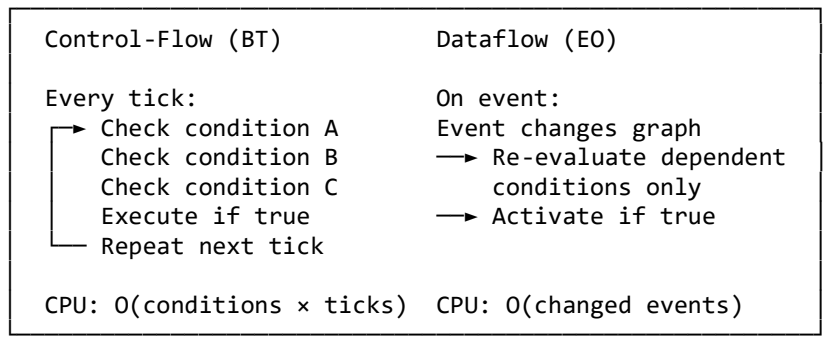

*Figure 1. Dataflow Execution vs Control-Flow.*

For game scenarios with many idle agents (open worlds, city NPCs, sleeping creatures), this difference is substantial. A BT checks every agent every tick. EO only processes agents whose state or the context influencing that state has changed.

---

[3] A formal justification for the executability of algorithms in the event-based dataflow paradigm is provided in Boldachev (2025b, Section 2).

## 3.4. Condition, SetValue, SetDo: The Logic Primitives

EO expresses all logic through three restriction types attached to model events:

*Condition* - a Boolean expression determining when an event can occur:

```
:: Condition: $.warmthLow == 1 && $.hasWood == 0 && $($.location).hasTree == 1
```

The action "gather wood" is available only when the agent is cold, has no wood, and is at a location with trees. If any part becomes false, the action becomes unavailable - no explicit "disable" needed.

*SetValue* - automatically computes a value when dependencies change:

```
:: SetValue: +$.warmth < +$.warmthMin
```

The `warmthLow` indicator is not set by game logic - it is *derived* from the relationship between `warmth` and `warmthMin`. Change either, and `warmthLow` updates automatically.

*SetDo* - triggers system actions when conditions are met:

```
:: SetDo: ({'$do': 'EditIndividual', '$IndividualID': $CurrentIndividual,
         '$Condition': $Value === "1", 'hasWood': 1})
```

When the player clicks "Gather Wood" (setting the trigger to "1"), the system automatically updates `hasWood`. The game logic is declarative - "gathering wood results in having wood" - not procedural.

## 3.5. Priority Through Condition Structure, Not Explicit Preemption

In BT, priority is structural: left branches execute before right branches. To make "warming" higher priority than "feeding," the designer places the warming subtree to the left.

In EO, priority emerges from condition logic. Consider:

```
# High priority: warming actions
: Attribute: action_gather
:: Condition: $.warmthLow == 1 && ...

# Lower priority: feeding actions
: Attribute: action_hunt
:: Condition: $.warmthLow == 0 && $.energyLow == 1 && ...
```

The hunt action *requires* `warmthLow == 0`. If warmth drops mid-hunt, this condition becomes false, and hunting is no longer available. No explicit interruption - the semantic state simply no longer permits that action.

This is not a trick or workaround; it reflects the domain truth: "You cannot hunt effectively while freezing." The model encodes why the priority exists, not just that it exists.

# 4. Practical Example: The Winter Feast Scenario

## 4.1. Scenario Description

To demonstrate EO principles in a game-relevant context, we implement a survival scenario where an agent must maintain two resources - warmth and energy - through a chain of dependent actions.

*Goal state.* Agent is "safe" when both `warmthLow == 0` and `energyLow == 0`.

Available actions:

- *Gather Wood* - collect wood from trees (requires: cold, no wood, trees present)
- *Light Fire* - create fire at location (requires: cold, has wood, no fire)
- *Hunt Deer* - obtain raw meat (requires: warm, hungry, no meat, deer present)
- *Cook Meat* - prepare food (requires: hungry, has raw meat, fire exists)
- *Eat Food* - restore energy (requires: hungry, has cooked meat)

*Critical design requirement.* Warmth has priority over energy. If the agent becomes cold during any feeding-related activity, warming actions must become available and feeding actions must become unavailable.

## 4.2. Model Implementation

The complete EO model defines concepts, properties, and the Survivor model with all conditions[4]:

```
Survivor: Model: Model Survivor
: Relation: location

# Editable states (for demo manipulation)
: Attribute: energy
: Attribute: warmth

# Configurable thresholds
: Attribute: energyMin
:: Default: 30
: Attribute: warmthMin
:: Default: 30

# Computed state indicators
: Attribute: energyLow
:: SetValue: +$.energy < +$.energyMin
: Attribute: warmthLow
:: SetValue: +$.warmth < +$.warmthMin
```

---

[4] Fully functional models are written in the Boldsea Semantic Language and interpreted (without compilation into code) by the boldsea framework.

```
# Inventory
: Attribute: hasWood
: Attribute: hasRawMeat
: Attribute: hasCookedMeat

# Goal state
: Attribute: isSafe
:: SetValue: $.energyLow == 0 && $.warmthLow == 0
```

## 4.3. Priority Actions: Warming (Priority 1)

Warming actions are available only when `warmthLow == 1`:

```
# GATHER WOOD
: Attribute: action_gather
:: Condition: $.warmthLow == 1 && $.hasWood == 0 && $($.location).hasTree == 1
:: SetDo: $.hasWood <- 1

# LIGHT FIRE
: Attribute: action_light_fire
:: Condition: $.warmthLow == 1 && $.hasWood == 1 && $($.location).hasFire == 0
:: SetDo: $.hasFire <- 1

# AUTOMATIC WARMING (reaction when fire exists)
: Attribute: _reaction_warm_up
:: SetValue: $($.location).hasFire == 1 && $.hasWood == 1 && $.warmthLow == 1
:: SetDo: $.warmth <- 70

# Note: for brevity, the system acts SetDo are presented in a simplified form.
# The full syntax is provided in Appendix A.
```

Note _reaction_warm_up: this is not a player action but an automatic reaction. When fire exists and the agent has wood and is cold, warming occurs automatically. The wood is consumed; warmth is restored.

## 4.4. Secondary Actions: Feeding (Priority 2)

Feeding actions require `warmthLow == 0` - the agent must be warm before pursuing food:

```
# HUNT DEER
: Attribute: action_hunt
:: Condition: $.warmthLow == 0 && $.energyLow == 1 && $.hasRawMeat == 0
            && $($.location).hasDeer == 1 && $.hasCookedMeat == 0
:: SetDo: $.hasRawMeat <- 1

# COOK MEAT
: Attribute: action_cook
:: Condition: $.energyLow == 1 && $.hasRawMeat == 1 && $.hasCookedMeat == 0
            && $($.location).hasFire == 1
:: SetDo: $.hasCookedMeat <- 1
```

```
# EAT FOOD
: Attribute: action_eat
:: Condition: $.energyLow == 1 && $.hasCookedMeat == 1
:: SetDo: $.energy <- 70
```

## 4.5. Emergent Behavior Through Condition Composition

The *Winter Feast* model demonstrates how complex behavioral sequences emerge from declarative conditions without explicit sequencing logic.

*Cascading dependencies.* Consider an agent who is hungry (`energyLow == 1`) and cold (`warmthLow == 1`). The goal is to eat, but examining action conditions reveals a dependency chain:

- *Eat* requires cooked meat → none available
- *Cook* requires raw meat and fire → no raw meat
- *Hunt* requires `warmthLow == 0` → agent is freezing, action blocked
- *Light fire* requires wood → no wood
- *Gather wood* → the only available action

No planning algorithm computed this sequence. The agent - or a simple heuristic selecting any available action - begins with "Gather Wood" because it is the sole unblocked option. Execution produces a "`hasWood`" event, which enables "`Light Fire`." Fire triggers the automatic warming reaction, setting `warmthLow` to false, which unblocks "`Hunt`." Each action's completion changes the graph state, causing the engine to re-evaluate dependent conditions and reveal the next available action.

*Figure 2. Winter Feast scenario interface. Button availability reflects current Condition state: with both `energyLow=1` and `warmthLow=1`, only "Gather Wood" is active - feeding actions are blocked by the warmth priority condition. An active button indicates the action is available (its Condition is satisfied); clicking the button signals action completion.*

The entire chain exists declaratively in the conditions. Behavior emerges from the interaction of rules, not from coded sequences or goal trees.

*Relation navigation.* The condition `$($.location).hasDeer == 1` demonstrates semantic graph traversal. Rather than querying a global registry or maintaining explicit references, the agent accesses location properties through its current `location` relation. If the agent moves to a different location, this check automatically yields a different result - hunting becomes available or unavailable based on the new environment, with no additional logic required.

*Automatic reactions vs. deliberate actions.* The _reaction_warm_up attribute models reflexive behavior - warming by a fire when cold. Unlike explicit actions requiring agent decision or button click, reactions trigger immediately when their SetValue condition becomes true. The agent does not "choose" to warm up; warming occurs as a natural consequence of being cold, having wood, and standing near fire. This distinction allows models to separate conscious decisions from automatic physical or physiological responses.

### 4.6. Demonstration: Priority Interruption

The following sequence demonstrates emergent priority behavior:

| Step | Manual Action | State Change | Available Actions | Principle Demonstrated |
| --- | --- | --- | --- | --- |
| 1 | Set energy → 20 | energyLow = 1 | Hunt Deer | Condition activation |
| 2 | Click "Hunt Deer" | hasRawMeat = 1 | Cook Meat (needs fire) | Action chaining |
| 3 | Set warmth → 20 | warmthLow = 1 | **Gather Wood** (Hunt disabled) | Priority interruption |
| 4 | Click "Gather Wood" | hasWood = 1 | Light Fire | Chain continues |
| 5 | Click "Light Fire" | hasFire = 1, warmth → 70 | Cook Meat | Warming complete, return to feeding |
| 6 | Click "Cook Meat" | hasCookedMeat = 1 | Eat Food | Resume interrupted chain |
| 7 | Click "Eat Food" | energy → 70, isSafe = 1 | None (goal reached) | Goal satisfaction |

*Table 2. Emergent priority behavior.*

*Critical observation.* At step 3, no "interrupt" command was issued. The hunt action simply became unavailable because its condition (warmthLow == 0) was no longer true. The system did not "decide" to prioritize warming - the semantic state made hunting impossible.

### 4.7. Comparison with BT Implementation

The BT encodes the same behavior, but:

- Priority is *structural* (left vs. right), not *semantic* (warmth requirement)
- Adding a third priority level requires restructuring the tree
- The reason for priority (survival requires warmth first) is implicit in position, not explicit in conditions
- In EO, adding a new priority (e.g., "flee from predator" with highest priority) requires only adding conditions that include the new requirement. Existing actions automatically become unavailable when the new priority condition is active.

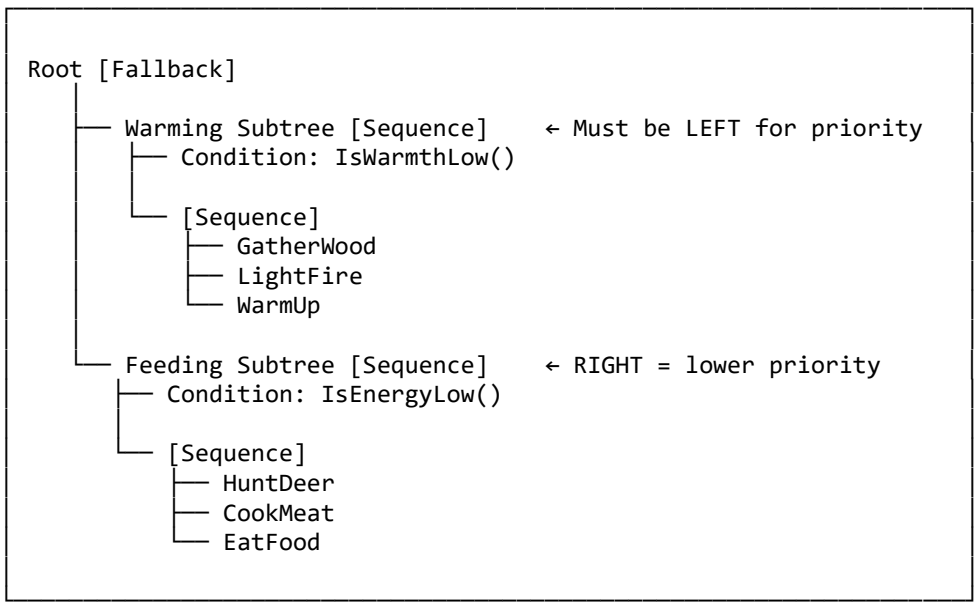

*Figure 3. An equivalent Behavior Trees structure for priority.*

## 4.8. Comparison with GOAP Implementation

A GOAP implementation would define a WorldState with predicates: `warmthLow`, `energyLow`, `hasWood`, `hasFire`, `hasRawMeat`, `hasCookedMeat`. Actions would be described through preconditions and effects: "gather" requires trees at location, produces `hasWood = true`; "light fire" requires `hasWood`, produces `hasFire = true`.

Priority would be implemented through cost functions or multiple goals: goal "stay warm" with weight 10, goal "stay fed" with weight 5 - the planner selects the shortest weighted path to satisfy the highest-priority unsatisfied goal.

The fundamental problem in many GOAP implementations is *state synchronization*. Who updates `warmthLow` in WorldState when warmth drops? A common pattern is to have a separate component poll the agent's warmth value each frame (or at fixed intervals), compare it against a threshold, and update the predicate. This tends to duplicate logic: the threshold check exists both as a computation (equivalent to `SetValue`) and as WorldState update code. If an update is forgotten or triggered under the wrong conditions, the planner may "believe" the agent is warm while the agent is actually freezing, producing erratic behavior.

In EO, warmthLow is not a cached predicate requiring synchronization - it is a *derived value* computed from the semantic graph. When warmth changes, any condition referencing `warmthLow` is re-evaluated through the subscription mechanism. There is no separate WorldState that must be kept in sync and no polling interval to tune; the semantic model *is* the world state. This greatly reduces the risk of manual synchronization bugs, even though application-level mistakes are, of course, still possible.

# 5. Integration Strategies

## 5.1. EO as High-Level Semantic Layer

EO is not intended to replace physics engines, animation systems, or low-level AI utilities. Instead, it serves as a high-level semantic layer that orchestrates these systems through a clear separation of concerns.

The semantic layer - implemented through EO - handles domain modeling, business rules, state computation, and decision availability. It operates at moderate frequency (10-20 Hz), activating only when relevant events occur. This layer determines *what* should happen: which actions are available, what state changes follow from completed actions, whether quest conditions are satisfied.

The execution layer - implemented through traditional ECS, behavior trees, or custom systems - handles animation state machines, pathfinding, physics simulation, and low-level AI such as steering and aiming. It operates at high frequency (60+ Hz) with deterministic per-frame costs. This layer determines *how* things happen: movement trajectories, animation blending, collision responses.

The temporal graph serves as the integration point. The execution layer reports completed actions as events; the semantic layer evaluates conditions and determines consequences. Commands flow downward (semantic → execution), events flow upward (execution → semantic → graph). This separation allows each layer to use appropriate technologies: ECS for cache-friendly batch processing, EO for semantic reasoning.

## 5.2. Beyond Tactical AI: Quest Systems

The WINTER_FEAST scenario addresses tactical decision-making: maintaining resources through immediate actions executed within minutes. Game quests operate at a different temporal scale - strategic goals spanning hours of gameplay, decomposed into stages with completion conditions. Traditional architectures treat these as separate systems: GOAP for moment-to-moment AI, dedicated quest frameworks for narrative progression.

EO reveals their fundamental unity. A quest stage is structurally identical to a GOAP action: both have preconditions (when is this available?) and effects (what changes upon completion?). The difference is granularity, not mechanism. Extending *Winter Feast* to a "Survive the Winter" quest illustrates this:

```
  : Attribute: day1_complete
  :: Condition: $.isSafe == 1
  :: SetValue: $.hours_passed >= 24

  : Attribute: day2_complete
  :: Condition: $.day1_complete == 1 && $.isSafe == 1
  :: SetValue: $.hours_passed >= 48
```

Each stage depends on the previous through condition chaining. The requirement `isSafe == 1` means the agent must maintain safe resource levels throughout - the engine tracks this automatically through dataflow, without explicit stage-completion code.

Quest composition emerges naturally. A main quest requiring shelter checks for a ShelterBuilt event; if absent, a subquest to build shelter becomes available. The main quest need not "know" about the subquest explicitly - it only verifies whether the required event exists in the graph. Subquest completion produces the event, satisfying the main quest's condition through normal dataflow propagation.

This unification eliminates adapter layers between tactical and strategic systems. Both GOAP actions and quest stages are models with conditions, validated against the same event graph. The implications for procedural content generation are significant: quests authored by designers and quests generated by LLMs use identical declarative format, validated by the same engine mechanisms.

### 5.3. Non-Invasive Extensibility

Adding new mechanics in traditional systems often requires modifying existing code due to tight coupling between components. A new component must integrate through explicit calls, shared state updates, and event subscriptions - violating encapsulation and creating risk of cascading failures.

EO provides non-invasive extensibility through a key property we can call *monotonicity of extension* at the level of model constraints. The dataflow architecture ensures that adding new models does not invalidate the *structural* correctness of the existing ontology: reification events are still created only if they satisfy their models. A new model event either integrates successfully - creating additional valid events when its conditions are met - or simply never activates. New models can, of course, change overall system behavior by introducing further events, but they do so via explicit conditions and effects rather than through hidden side effects on shared mutable state.

New mechanics are defined as semantic models, added to the ontology, and immediately begin reacting to existing events and generating new ones. Existing models remain unmodified; they simply gain access to new event types if their conditions reference them. This enables modular content architecture: a base game provides core mechanics, while extensions add new interaction types declaratively - magic systems, crafting, diplomacy - all operating through the unified event graph without explicit integration code.

### 5.4. Validation and Static Analysis

The declarative nature of EO enables automatic correctness analysis before execution. Traditional imperative code can be checked syntactically by compilers, but logical errors - incorrect sequences, forgotten edge cases, violated invariants - manifest only at runtime.

EO permits static verification of semantic properties. The system builds a dependency graph between models by analyzing conditions and effects. If model A generates event type E, and model B requires E in its condition, a dependency A → B exists. Analyzing this graph at load time detects several classes of problems.

*Unreachable models* - those whose conditions can never be satisfied - are discovered automatically. If a condition references an event type that no other model can generate, the system warns: "Model X requires event Y, but Y cannot occur." The designer fixes the problem before players encounter an impossible quest.

*Type safety* is enforced by checking attribute references in conditions. If a condition check `$($.location).hasDeer` but the location model has no `hasDeer` attribute, this is caught at load time, not runtime.

These checks are particularly valuable for procedurally generated content. An LLM may generate semantically reasonable but syntactically incorrect models. Static validation detects errors and returns detailed descriptions, enabling iterative refinement until a valid result is achieved.

In the current boldsea prototypes, static analysis focuses on these local properties: type checking of attribute and relation references, basic dependency graph construction, and simple reachability checks based on declared event types. More ambitious analyses - such as proving global invariants, exploring entire spaces of event traces, or integrating with external model checkers - are not part of the present implementation and should be viewed as directions for future work rather than existing features.

## 5.5. Debugging and Observability

In control-flow systems, logic errors often manifest as hangs, infinite loops, or stuck states. In EO, this error category is greatly reduced and takes a different form. A model event either has its condition satisfied and can activate, or it does not - in both cases the engine continues processing other events. If a condition is never satisfied, the action is simply never available. The "error" is a design issue (incorrect condition), not a runtime failure.

*Temporal traceability* is native to the architecture. Every event contains its causal history: Actor (who created it), Cause (which events were prerequisites), Model (which template validated it). To debug "why does the agent have no wood?", traverse the graph: find the most recent `hasWood` event, trace what set it to zero (the warming reaction consumed it), trace what triggered that reaction. This is graph traversal, not log parsing - the debugging information *is* the data structure.

*Reproducibility* follows directly. To reproduce a bug in BT requires capturing exact blackboard state, tree position, and external system states. In EO: export the relevant subgraph, load into a fresh engine - behavior is deterministic from graph state. This enables bug reports as graph exports, unit tests as graph fixtures, and regression testing through graph comparison.

*Multiperspectivity* maps to game-specific debugging. Fog of war: NPC A believes the player is in Room 1 while NPC B knows Room 2 - both exist as events from different actors. Illusions: NPCs reacting to a decoy have causal chains tracing to decoy-actor events. Client-server disagreement: both "hit landed" and "hit missed" events exist; reconciliation logic determines the authoritative outcome. Debugging means inspecting which actor's events influenced which decisions - information native to the graph.

# 6. Comparative Analysis BT, GOAP and EO

## 6.1. Architectural Comparison

The following table summarizes the fundamental architectural differences between BT, GOAP, and EO:

| Aspect | Behavior Trees | GOAP | Executable Ontologies |
| --- | --- | --- | --- |
| Unit of control | Node (command) | Action (operator) | Model Event (condition) |
| Execution model | Tick-based polling | Plan-then-execute | Event-driven subscription |
| State representation | External blackboard | Flat world state | Temporal semantic graph |
| Priority mechanism | Structural (tree position) | Goal utility ranking | Condition logic |
| Interruption | Explicit preemption | Replan on failure | Condition becomes false |
| History | None (volatile) | None (snapshots) | Native (immutable graph) |
| Runtime modification | Requires restart | Requires restart | Zero-downtime |
| Validation | External testing | Precondition checking | Model-based (continuous) |

*Table 4. Architectural differences between BT, GOAP and EO*

## 6.2. Reactivity Mechanisms

All three approaches achieve reactivity, but through fundamentally different mechanisms:

*BT: Polling-based reactivity.* Every tick, the tree is traversed from root. High-priority branches (left) are checked before low-priority (right). If a high-priority condition becomes true, its subtree executes, preempting lower branches. Cost: O(nodes × ticks).

*GOAP: Replan-based reactivity.* The planner generates a sequence of actions. During execution, if a precondition fails, the system replans. Reactivity depends on how frequently the planner is invoked and how fast it can generate new plans. Cost: O(replan frequency × search complexity).

*EO: Subscription-based reactivity.* When an event changes the graph, only dependent conditions are re-evaluated. If warmth changes, only conditions referencing warmth or `warmthLow` are checked. No global traversal. Cost: O(affected conditions).

For scenarios with many idle agents (city NPCs, wildlife, dormant enemies), EO's subscription model provides significant efficiency gains. An agent standing idle generates zero CPU cost until something relevant changes in its semantic neighborhood.

### 6.3. Modularity and Extension

*Adding a new action in each paradigm:*

*BT:* Create new leaf nodes, wire them into the tree structure, ensure correct priority positioning. If the new action has higher priority than existing actions, tree restructuring may be required.

*GOAP:* Define new action with preconditions and effects. The planner automatically incorporates it into plans where applicable. However, debugging emergent plans can be challenging.

*EO:* Add a new model event with its Condition. No structural changes to existing events. The new action becomes available when its condition is true, independent of other actions.

*Adding a new priority level:*

*BT:* Restructure tree - insert new subtree at appropriate position, potentially rewiring existing branches.

*GOAP:* Modify goal utility function or add new goal with higher priority. Plan generation adapts, but interactions may be unpredictable.

*EO:* Add condition clause to lower-priority actions (e.g., `&& predatorNearby == 0`). No structural changes. Lower-priority actions automatically become unavailable when the new condition applies.

## 7. Limitations and Future Work

### 7.1. Adoption Challenges

The most significant barrier to EO adoption is conceptual rather than technical. Developers trained in imperative programming think in sequences: "first do X, then check Y, then do Z." EO requires thinking in conditions: "X is possible when A and B are true." This shift - from specifying *what to do* to declaring *when it's possible* - demands retraining intuitions built over years of procedural practice.

The Winter Feast scenario appears simple in retrospect, but designers accustomed to scripting behavior trees may initially struggle to express intent declaratively. The question "how do I make the agent gather wood?" has no direct answer in EO; instead, one must ask "under what world conditions is wood-gathering possible?" This inversion is powerful once internalized but creates a learning curve.

As a new technology, EO also lacks the ecosystem that established approaches enjoy. There are no extensive libraries of reusable models, no integration plugins for major engines like Unity or Unreal, and a small community producing documentation and tutorials. This bootstrapping problem - adoption requires ecosystem, ecosystem requires adoption - is common to architecture shifts and will resolve only through demonstrated value in production systems.

### 7.2. Performance Considerations

While EO's subscription model is efficient for sparse activity - thousands of idle NPCs generate zero CPU cost - certain scenarios present challenges. Mass simultaneous events, such as an explosion affecting hundreds of agents, trigger extensive condition re-evaluation. Unlike behavior trees' predictable per-tick cost, EO can exhibit load spikes when many graph changes cascade through dependent conditions.

Deep condition chains referencing multiple nested properties require graph traversals that can become expensive in pathological cases. Long-running sessions accumulate events, and without active management, query performance degrades as the graph grows.

These challenges have known mitigations. Hybrid architecture assigns high-frequency systems (physics, damage calculation) to traditional ECS while reserving EO for semantic reasoning. Model indexing limits condition evaluation to relevant models. Graph partitioning isolates subgraphs by region or actor group. Transitive reduction compresses completed action chains. However, systematic benchmarking under game-realistic loads - thousands of concurrent agents, complex interlocking quest systems, extended play sessions - remains an area requiring empirical investigation.

Criticism of event-driven architectures often focuses on propagation overhead. In systems with high event frequency - physics collisions, animation triggers - each event potentially activates many models for condition checking, creating load spikes.

*Model indexing* addresses this directly. Instead of checking all model conditions on every event, the engine builds an index: for each event type, a list of models whose conditions reference that type. When event E occurs, only indexed models are checked - reducing evaluations by orders of magnitude in large ontologies. The index is built once at load time and updated only when models are added.

*Hybrid architecture* separates systems by update frequency. Low-level systems requiring high frequency and deterministic performance - physics, rendering, animation - use traditional approaches or ECS. High-level logic - AI decisions, quests, world simulation - uses EO. Integration occurs through aggregated events: the physics system reports "collision occurred" as a single event; EO determines the semantic consequences.

*Memory optimization* through transitive reduction compresses completed action chains. A crafting sequence (`CheckResources → ReserveTable → StartAnimation → ... → ItemCreated`) reduces to (`Intent:Craft → Result:ItemCreated`). Intermediate events are archived for debugging but removed from the active graph. The causal relationship is preserved; queries remain efficient. This aligns with save systems: files store high-level facts, not micro-events.

*Graph partitioning* handles long-running sessions. Events are partitioned by owner (character, location, guild) and time window. An "active frontier" of recent events remains in memory; older events archive to database with reference stubs for recovery. Critical events - quest acceptance, achievements, story choices - are marked persistent and excluded from archiving. The system analyzes model conditions to determine which event types require persistence.

### 7.3. Future Directions: LLM Integration

The declarative nature of EO creates unique opportunities for integration with Large Language Models across multiple dimensions. At present, these ideas are explored only in small-scale experiments; they are not part of a production-ready boldsea toolchain.

*Model generation* allows designers to describe behavior in natural language while LLMs produce formal EO models. A description like "wolves hunt in packs - a lone wolf flees from humans, but a pack of three or more attacks" can be transformed into conditions and computed attributes. Because models are validated at load time, generation errors are caught immediately rather than manifesting as runtime failures. This enables iterative refinement: the LLM generates, the engine validates, errors are fed back, the LLM corrects.

*Narrative generation* leverages the temporal graph as a source of ground truth. The graph contains complete causal history - what happened and why. An LLM can traverse relevant events and generate natural language descriptions that reflect actual player or NPC experience rather than scripted templates. The survivor who struggled through bitter cold, barely managed to gather wood, and finally lit a fire has a story embedded in the event graph; the LLM merely translates it to prose.

*Runtime model adaptation* represents the most experimental direction. When an agent reaches a dead end - no actions available, goal unachieved - an LLM could analyze the semantic context and propose model extensions. An agent needing to cross a river with no bridge might receive a dynamically generated "build raft" action if trees and tools are available. This runtime ontology extension expands the game's possibility space based on context, though ensuring coherence and balance in such generative systems requires further research.

## 7.4. Multiplayer and Distributed Systems

EO's rejection of global time - ordering events by causality rather than timestamps - suggests interesting implications for networked games. Traditional MMO architectures suffer from synchronization problems: conflicts during simultaneous player actions, complex rollback procedures when detecting cheating, and authoritative servers creating bottlenecks.

Causal event graphs suggest an alternative approach. Each client generates events with references to causally prior events. Servers merge event streams through causal analysis rather than timestamp comparison. Conflict resolution examines causal chains to determine which events have valid prerequisites, eliminating entire classes of race conditions inherent to timestamp-based systems.

This architecture also offers natural approaches to lag compensation. "What the player saw" is recorded as events from the player's actor perspective; reconciliation analyzes causal validity rather than requiring timestamp rollback. Sharding becomes straightforward: subgraphs partition across servers with well-defined merge semantics based on causal dependencies.

These properties suggest EO may be particularly suited to next-generation MMO architectures, but practical validation at scale is needed. Key research questions include efficient algorithms for distributed consensus over event graphs, serialization strategies for network transmission, and conflict resolution policies that maintain gameplay fairness while preserving causal integrity.

These ideas have so far been investigated only at the level of architectural sketches and small prototypes; validating them at MMO scale remains future work.

### 7.5. Formal Verification and Machine Learning

Two additional research directions warrant mention. They are not implemented in the current boldsea engine, but follow naturally from its declarative semantics. Model checking could prove properties of game systems: all quests are achievable from the starting state, no dialogue trees contain deadlocks, resource economies cannot be exploited through unintended action sequences. Adapting existing verification techniques to event graph semantics and developing domain-specific specification languages for game properties represent promising theoretical work.

Machine learning offers a complementary direction. While the structure of behavior is defined declaratively by designers, numerical parameters - thresholds, weights, costs - could be optimized through reinforcement learning. An agent with ontology-defined behavior trains in simulation; the learning algorithm adjusts parameters to maximize reward. This combines the interpretability of declarative rules with the power of adaptive optimization, avoiding the black-box nature of purely neural approaches while gaining the benefits of empirical tuning (Menemencioğlu, 2025).

# 8. Conclusion

This section discusses limitations and prospective directions for EO as realized in the current boldsea prototypes. The mechanisms outlined here - large-scale performance tuning, deep engine integration, MMO architectures, formal verification pipelines, and tight LLM integration - should be understood primarily as research and engineering directions. The existing implementation provides the core event-graph semantics, dataflow execution, and local static validation; the rest of this section outlines how these foundations could be extended.

### 8.1. Summary of Contributions

*Conceptual reframing.* We have argued that EO represents not merely a different technique, but a different level of abstraction. BT and GOAP model *behavior* - what agents should do. EO models *the world* - what entities exist, what properties they have, under what conditions events can occur. Behavior emerges as a consequence of semantic structure rather than being explicitly programmed.

*Practical demonstration.* The Winter Feast scenario illustrated how priority-based task interruption arises naturally from condition logic. No preemption mechanism was coded; the warming actions became available and feeding actions became unavailable as a direct consequence of the `warmthLow` state changing. The "intelligence" is in the ontology, not in control flow.

*Integration analysis.* We have shown that EO need not replace existing systems but can serve as a high-level semantic layer, providing decision availability and state computation while execution-level systems handle animation, physics, and pathfinding.

### 8.2. The Paradigm Shift

The transition from BT to EO mirrors historical shifts in software engineering:

| Era | Data | Logic | Characteristic |
| --- | --- | --- | --- |
| Procedural | Global variables | Scattered in procedures | "Go to" thinking |
| Object-Oriented | Encapsulated in objects | Methods on objects | "Tell, don't ask" |
| Semantic | Unified in graph | Declarative conditions | "Model the world" |

*Table 5. Historical shifts in software engineering.*

Just as OOP unified data and behavior in objects, EO unifies data, logic, and rules in semantic models. The question changes from "How do I program this agent?" to "What is true about this world?"

## 8.3. Recommendations

*For researchers.* The intersection of executable ontologies with adjacent fields offers rich opportunities. Formal verification techniques could prove game invariants - quest achievability, dialogue completeness, economy balance. Distributed consensus algorithms for causal event graphs could enable new MMO architectures. Reinforcement learning for parameter optimization combines interpretable rules with adaptive tuning. Empirical studies comparing designer productivity across paradigms would provide practical guidance.

*For game developers.* Consider EO for systems where semantic clarity matters more than raw performance: quest systems, NPC social behavior, world state rules, faction relationships. The potential for LLM-driven content generation - quests, dialogues, dynamic mechanics - makes EO particularly attractive for games requiring extensive authored or procedural content. Hybrid architectures combining EO with ECS or behavior trees may offer the best of both paradigms.

*For tool developers.* The ecosystem gap is real and represents opportunity. Integration plugins for Unity and Unreal, visual model editors with validation feedback, and debugging tools for causal graph inspection would significantly lower adoption barriers and accelerate the technology's maturation.

## 8.4. Closing Perspective

Games are, fundamentally, simulated worlds with rules. The more complex and dynamic these worlds become, the more the semantic-process gap constrains development. Executable Ontologies offer a path toward closing this gap - not by writing better algorithms, but by modeling worlds more truly.

The agent that gathers wood when cold does not "decide" to gather wood. The world is simply in a state where gathering wood is possible and eating is not. The behavior is not programmed; it is *entailed*.

This is the promise of executable ontologies: games where behavior emerges from meaning, and where the designer's intent - expressed as declarative knowledge - becomes directly executable reality.

# Appendix A. Scenarios Code

We present the comprehensive EO code that implements the Winter Feast Scenarios. The code, as presented, is loaded into the graph via the engine's console. Upon loading, it undergoes full validation for both format and semantic consistency against the previously loaded code (ensuring the presence of required concepts, properties, model events, and individuals).

For brevity and clarity, the code does not include the Permission restricting properties, which manage access rights. The code for SetDo system acts is provided in full, rather than as pseudocode, as seen in the examples in the main text.

User Interface (UI) pages are described in BSL as individuals of the View concept models and are interpreted by the View-controller. The properties of the View-models are not listed, as they are part of the genesis block of system events in the graph, which is loaded when the engine starts.

```
# SURVIVAL SCENARIO: WINTER FEAST

# CONCEPTS
Concept: Instance: Survivor
Concept: Instance: Location

# PROPERTIES
Attribute: Individual: energy
: DataType: Numeric
Attribute: Individual: warmth
: DataType: Numeric
Attribute: Individual: energyMin
: DataType: Numeric
Attribute: Individual: warmthMin
: DataType: Numeric
Attribute: Individual: energyLow
: DataType: Boolean
Attribute: Individual: warmthLow
: DataType: Boolean
Attribute: Individual: hasWood
: DataType: Boolean
Attribute: Individual: hasRawMeat
: DataType: Boolean
Attribute: Individual: hasCookedMeat
: DataType: Boolean
Attribute: Individual: _reaction_warm_up
: DataType: Boolean
Attribute: Individual: hasTree
: DataType: Boolean
Attribute: Individual: hasDeer
: DataType: Boolean
Attribute: Individual: hasFire
: DataType: Boolean
Relation: Individual: location
: Range: Location
Relation: Individual: survivor
: Range: Survivor
Attribute: Individual: action_gather
: DataType: Boolean
Attribute: Individual: action_light_fire
: DataType: Boolean
Attribute: Individual: action_hunt
: DataType: Boolean
Attribute: Individual: action_cook
: DataType: Boolean
Attribute: Individual: action_eat
: DataType: Boolean
Attribute: Individual: isSafe
: DataType: Boolean

# MODELS
Location: Model: Model Location
: Relation: survivor
: Attribute: hasTree
: Attribute: hasDeer
: Attribute: hasFire

Survivor: Model: Model Survivor
: Relation: location
: Attribute: energy
: Attribute: warmth
: Attribute: energyMin
:: Default: 30
: Attribute: warmthMin
:: Default: 30

: Attribute: energyLow
:: SetValue: +$.energy < +$.energyMin
: Attribute: warmthLow
:: SetValue: +$.warmth < +$.warmthMin
: Attribute: hasWood
: Attribute: hasRawMeat
: Attribute: hasCookedMeat
: Attribute: isSafe
:: SetValue: $.energyLow == 0
   && $.warmthLow == 0

: Attribute: action_gather
:: Condition: $.warmthLow == 1
   && $.hasWood == 0
   && $($.location).hasTree == 1
:: SetDo: ({'$do': 'EditIndividual',
   '$IndividualID': $CurrentIndividual,
   '$Condition': $Value === "1",'hasWood': 1})
: Attribute: action_light_fire
:: Condition: $.warmthLow == 1
   && $.hasWood == 1
   && $($.location).hasFire == 0
```

```
:: SetDo: ({'$do': 'EditIndividual',
   '$IndividualID': $.location, '$Condition':
   $Value === "1", 'hasFire': 1})

: Attribute: _reaction_warm_up
:: SetValue: $($.location).hasFire == 1
   && $.hasWood == 1 && $.warmthLow == 1
:: SetDo: ({'$do': 'EditIndividual',
   '$IndividualID': $CurrentIndividual,
   '$Condition': $Value === "1",
   'hasWood': 0, 'warmth': 70})

: Attribute: action_hunt
:: Condition: $.warmthLow == 0
   && $.energyLow == 1 && $.hasRawMeat == 0
   && $($.location).hasDeer == 1
   && $.hasCookedMeat == 0
:: SetDo: ({'$do': 'EditIndividual',
   '$IndividualID': $CurrentIndividual,
   '$Condition': $Value === "1",
   'hasRawMeat': 1})

: Attribute: action_cook
:: Condition: $.energyLow == 1
   && $.hasRawMeat == 1
   && $.hasCookedMeat == 0
   && $($.location).hasFire == 1
:: SetDo: ({'$do': 'EditIndividual',
   '$IndividualID': $CurrentIndividual,
   '$Condition': $Value === "1",
   'hasRawMeat': 0, 'hasCookedMeat': 1})

: Attribute: action_eat
:: Condition: $.energyLow == 1
   && $.hasCookedMeat == 1
:: SetDo: ({'$do': 'EditIndividual',
   '$IndividualID': $CurrentIndividual,
   '$Condition': $Value === "1",
   'hasCookedMeat': 0, 'energy': 70})

# INDIVIDUALS
Location: Individual: Forest Clearing
: SetModel: Model Location
: survivor: John Doe
: hasTree: 1
: hasDeer: 1
: hasFire: 0

Survivor: Individual: John Doe
: SetModel: Model Survivor
: location: Forest Clearing
: energy: 50
: warmth: 50
: energyMin: 30
: warmthMin: 30
: hasWood: 0
: hasRawMeat: 0
: hasCookedMeat: 0

# VIEW LAYER
View: Model: Model View Individual
: Attribute: ConceptPage
: Relation: IndividualID
: Attribute: ViewConcept
:: Relation: IndividualList
::: SetValue: $.IndividualID
:: Attribute: ViewMode
:: Attribute: Title
:: Attribute: Include
::: Multiple: 1
:: Attribute: Exclude
::: Multiple: 1
:: Attribute: Control
::: Multiple: 1
::: Attribute: Title
::: Attribute: ControlType
::: Attribute: Value

View: Individual: View Survivor
: SetModel: Model View Individual
: ConceptPage: Survivor
: IndividualID: John Doe
: ViewConcept: Survivor
:: IndividualList: John Doe
:: ViewMode: showcase
:: Exclude: _reaction_warm_up
:: Exclude: energyMin
:: Exclude: warmthMin
:: Control: action_gather
::: Title: Gather Wood
::: ControlType: button
::: Value: 1
:: Control: action_light_fire
::: Title: Light Fire
::: ControlType: button
::: Value: 1
:: Control: action_hunt
::: Title: Hunt Deer
::: ControlType: button
::: Value: 1
:: Control: action_cook
::: Title: Cook Meat
::: ControlType: button
::: Value: 1
:: Control: action_eat
::: Title: Eat Food
::: ControlType: button
::: Value: 1

View: Individual: View Location
: SetModel: Model View Individual
: ConceptPage: Location
: IndividualID: Forest Clearing
: ViewConcept: Location
:: IndividualList: Forest Clearing
:: ViewMode: showcase
```

# Appendix B. Boldsea IDE Screenshots

*Figure 4. boldsea IDE, Temporal Graph.*

*Figure 5. boldsea IDE, Model Editor.*

*Figure 6. boldsea IDE, Restriction Editor.*

25